\documentclass{article}
\usepackage{icml2001,mlapa}
\usepackage{graphics,graphicx,amsmath}
\begin{document} 

\newcommand{\argmax}[1]{\underset{#1}{\rm argmax}\quad}
\newcommand{\argmin}[1]{\underset{#1}{\rm argmin}\quad}
\newcommand{\amax}[1]{\underset{#1}{\rm max}\quad}
\newcommand{\amin}[1]{\underset{#1}{\rm min}\quad}
\newcommand{\Lex}{{\bf L}}
\newcommand{\W}{{\bf W}}
\newcommand{\A}{{\bf A}}
\newcommand{\p}{{\rm P}}
\newcommand{\C}{{\bf C}}
\newcommand{\X}{{\bf X}}

\twocolumn[
\icmltitle{A procedure for unsupervised lexicon learning}

\icmlauthor{Anand Venkataraman}{anand@speech.sri.com}
\icmladdress{Speech Technology and Research  Lab\\
SRI International\\
333 Ravenswood Ave\\
Menlo Park, CA 94025}
\vskip 0.3in
]
 
\begin{abstract} 
We describe an incremental unsupervised procedure to learn words from
transcribed continuous speech.  The algorithm is based on a
conservative and traditional statistical model, and results of
empirical tests show that it is competitive with other algorithms that
have been proposed recently for this task.
\end{abstract} 

\section{Introduction}

English speech lacks the acoustic analog of blank spaces that people
are accustomed to seeing between words in written text.  Discovering
words in continuous spoken speech then is an interesting problem that
has been treated at length in the literature.  The issue is also
particularly prominent in the parsing of written text in languages
that do not explicitly include spaces between words.  

In this paper, we describe an incremental unsupervised algorithm based
on a formal statistical model to infer word boundaries from continuous
speech.  The main contributions of this study are as follows: First,
it demonstrates the applicability and competitiveness of a
conservative traditional approach for a task for which nontraditional
approaches have been proposed even recently
\cite{Brent:EPS99,Brent:DRP96,deMarcken:UAL95,Elman:FST90,Christiansen:LSS98}.
Second, although the model leads to the development of an algorithm
that learns the lexicon in an unsupervised fashion, results of partial
supervision are also presented, showing that its performance is
consistent with results from learning theory.

\section{Related Work}

While there exists a reasonable body of literature with regard to word
discovery and text segmentation, especially with respect to languages
such as Chinese and Japanese, which do not explicitly include spaces
between words, most of the statistically based models and algorithms
tend to fall into the supervised learning category.  These require the
model to first be trained on a large corpus of text before it can
segment its input.\footnote{See, for example, \emcite{Zimin98:CTS93} for
a survey and \emcite{Teahan:CBA00} for the most recent such approach.}
It is only of late that interest in unsupervised algorithms for text
segmentation seems to have gained ground.  In the last ANLP/NAACL
joint language technology conference, \emcite{Ando:MUS00} proposed an
algorithm to infer word boundaries from character $n$-gram statistics
of Japanese Kanji strings.  For example, a decision to insert a word
boundary between two characters is made solely based on whether
character $n$-grams adjacent to the proposed boundary are relatively
more frequent than character $n$-grams that straddle it.  However,
even this algorithm is not truly unsupervised.  There is a threshold
parameter involved that must be tuned in order to get optimal
segmentations when single character words are present.  Also, the
number of orders of $n$-grams that are significant in the segmentation
decision making process is a tunable parameter.  The authors state
that these parameters can be set with a very small number of
pre-segmented training examples, as a consequence of which they call
their algorithm {\em mostly unsupervised}.  A further factor
contributing to the incommensurability of their algorithm with our
approach is that it is not immediately obvious how to adapt their
algorithm to operate incrementally.  Their procedure is more suited to
batch segmentation, where corpus $n$-gram statistics can be obtained
during a first pass and segmentation decisions made during the second.
Our algorithm, however, is purely incremental and unsupervised and
does not need to make multiple passes over the data, nor require
tunable parameters to be set from training data beforehand.  In this
respect, it is most similar to Model Based Dynamic Programming,
hereafter referred to as MBDP-1, which has been proposed in
\cite{Brent:EPS99}.  To the author's knowledge, MBDP-1 is probably the
most recent and only other completely unsupervised work that attempts
to discover word boundaries from unsegmented speech data.  Both the
approach presented in this paper and MBDP-1 are based on explicit
probability models.  As the name implies, MBDP-1 uses dynamic
programming to infer the best segmentation of the input corpus.  It is
assumed that the entire input corpus, consisting of a concatenation of
all utterances in sequence, is a single event in probability space and
that the best segmentation of each utterance is implied by the best
segmentation of the corpus itself.  The model thus focuses on
explicitly calculating probabilities for every possible segmentation
of the entire corpus, subsequently picking the segmentation with the
maximum probability.  More precisely, the model attempts to calculate
$$
\p(\bar{w}_m) = \sum_n \sum_L \sum_f \sum_s
	\p(\bar{w}_m|n,L,f,s) \cdot \p(n,L,f,s)
$$
for each possible segmentation of the input corpus where the left-hand
side is the exact probability of that particular segmentation of the
corpus into words $\bar{w}_m = w_1 w_2 \cdots w_m$ and the sums are
over all possible numbers of words, $n$, in the lexicon, all possible
lexicons, $L$, all possible frequencies, $f$, of the individual words
in this lexicon and all possible orders of words, $s$, in the
segmentation.  In practice, the implementation uses an incremental
approach that computes the best segmentation of the entire corpus up
to step $i$, where the $i$th step is the corpus up to and including
the $i$th utterance.  Incremental performance is thus obtained by
computing this quantity anew after each segmentation $i-1$, assuming,
however, that segmentations of utterances up to but not including $i$
are fixed.  Thus, although the segmentation algorithm itself is
incremental, the formal statistical model of segmentation is not.

Furthermore, making the assumption that the corpus is a single event
in probability space significantly increases the computational
complexity of the incremental algorithm.  The approach presented in
this paper circumvents these problems through the use of a
conservative statistical model that is directly implementable as an
incremental algorithm.  In the following sections, we describe the
model and the algorithm derived from it.  The technique can basically
be seen as an modification of Brent's work, borrowing in particular
his successful dynamic programming approach while substituting his
statistical model with a more conservative one.

\section{Model Description}
\label{sec:model}

The language model described here is fairly standard in nature.  The
interested reader is referred to \emcite[p.57--78]{Jelinek:SMS97},
where a detailed exposition can be found.  Basically, we seek
\begin{eqnarray}
\label{eqn:w2}
\hat{\W} &=& \argmax{\W} \p(\W)\\
         &=& \argmax{\W} \prod_{i=1}^n \p(w_i|w_1, \cdots, w_{i-1})\\
         &=& \argmin{\W} \sum_{i=1}^n -\log \p(w_i|w_1, \cdots,
         w_{i-1}) \label{eqn:pw}
\end{eqnarray}
where $\W = w_1, \cdots, w_n$ denotes a particular string of $n$
words.  Each word is assumed to be made up of a finite sequence of
characters representing phonemes from a finite inventory.

We make the unigram approximation that word histories are irrelevant
to their probabilities.  This allows us to rewrite the right-hand side
of Equation~\ref{eqn:pw} as unconditional probabilities.  We also
employ back-off \cite{Katz:EPF87} using the Witten-Bell technique
\cite{Witten:ZFP91} when novel words are encountered.  This enables us
to use an open vocabulary and estimate familiar word probabilities
from their relative frequencies in the observed corpus while backing
off to the letter level for novel words.  In our case, a novel word is
decomposed into its constituent phonemes and its probability is then
calculated as the normalized product of its phoneme probabilities.  To
do this, we introduce the sentinel phoneme '\#', which is assumed to
terminate every word.  The model can now be summarized very simply as
follows:

\begin{eqnarray}
\p(w) &=& \left\{
	\begin{array}{ll}
	\frac{C(w_i)}{N + S} & {\rm if\ } C(w) > 0 \\
	\frac{N}{N + S} \p_\Sigma(w) & {\rm otherwise}\\
	\end{array}
	\right. \label{eqn:smooth1}\\
\p_\Sigma(w) &=& \frac{r(\#)\prod\limits_{j=1}^{|w|}
	r(w[j])}{1-r(\#)} \label{eqn:pnovel}
\end{eqnarray}
where $C()$ denotes the count or frequency function, $N$ denotes the
number of distinct words in the word table, $S$ denotes the sum of
their frequencies, $|w|$ denotes the length of word $w$, excluding the
sentinel `\#', $w[j]$ denotes its $j$th phoneme, and $r()$
denotes the relative frequency function.  The normalization by
dividing using $1-r(\#)$ in Equation~(\ref{eqn:pnovel}) is necessary
because otherwise
\begin{eqnarray}
\sum_w \p(w) &=& \sum_{i=1}^\infty (1-\p(\#))^i\p(\#)\\
&=& 1-\p(\#)
\end{eqnarray}
Since we estimate $\p(w[j])$ by $r(w[j])$, dividing by $1-r(\#)$ will ensure
that $\sum_w \p(w) = 1.$

\section{Method}

As in \emcite{Brent:EPS99}, the model described in
Section~\ref{sec:model} is presented as an incremental learner.  The
only knowledge built into the system at start-up is the phoneme table
with a uniform distribution over all phonemes, including the sentinel
phoneme.  The learning algorithm considers each utterance in turn and
computes the most probable segmentation of the utterance using a
Viterbi search \cite{Viterbi:EBC67} implemented as a dynamic
programming algorithm described shortly.  The most likely placement of
word boundaries computed thus is committed to before considering the
next presented utterance.  Committing to a segmentation involves
learning word probabilities as well as phoneme probabilities from the
inferred words.  These are used to update their respective tables.
To account for effects that any specific ordering of input utterances
may have on the segmentations that are output, the performance of the
algorithm is averaged over 1000 runs, with each run receiving as input
a random permutation of the input corpus.

\subsection*{The input corpus}

The corpus, which is identical to the one used by
\emcite{Brent:EPS99}, consists of orthographic transcripts made by
\emcite{Bernstein:PPC87} from the CHILDES collection
\cite{MACWHINNEY:CLD85}.  The speakers in this study were nine mothers
speaking freely to their children, whose ages averaged 18 months
(range 13--21).  Brent and his colleagues also transcribed the corpus
phonemically (using an ASCII phonemic representation), ensuring that
the number of subjective judgments in the pronunciation of words was
minimized by transcribing every occurrence of the same word
identically.  For example, ``look'', ``drink'' and ``doggie'' were
always transcribed ``lUk'', ``drINk'' and ``dOgi'' regardless of where
in the utterance they occurred and which mother uttered them in what
way.  Thus transcribed, the corpus consists of a total of 9790 such
utterances and 33,399 words including one space after each word and
one newline after each utterance.

It is noteworthy that the choice of this particular corpus for
experimentation is motivated purely by its use in
\emcite{Brent:EPS99}.  The algorithm is equally applicable to plain
text in English or other languages.  The main advantage of the CHILDES
corpus is that it allows for ready and quick comparison with results
hitherto obtained and reported in the literature.  Indeed, the
relative performance of all the discussed algorithms is mostly
unchanged when tested on the 1997 Switchboard telephone speech corpus
with disfluency events removed.

\section{Algorithm}
\newcommand{\seg}{{\bf seg}}
\newcommand{\word}{{\bf word}}

The dynamic programming algorithm finds the most probable word
sequence for each input utterance by assigning to each segmentation a
score equal to the logarithm of its probability and committing to the
segmentation with the highest score.  In practice, the implementation
computes the negative logarithm of this score and thus commits to the
segmentation with the least negative logarithm of the probability.
The algorithm is presented in recursive form in
Figure~\ref{fig:dyn-rec} for readability.  The actual implementation,
however, used an iterative version.  The algorithm to evaluate the
back-off probability of a word is given in Figure~\ref{fig:word-len}.
Essentially, the algorithm description can be summed up semiformally
as follows: For each input utterance $u$, which has either been read
in without spaces, or from which spaces have been deleted, we evaluate
every possible way of segmenting it as $u = u' + w$ where $u'$ is a
subutterance from the beginning of the original utterance up to some
point within it and $w$, the lexical difference between $u$ and $u'$,
is treated as a word.  The subutterance $u'$ is itself evaluated
recursively using the same algorithm.  The base case for recursion
when the algorithm rewinds is obtained when a subutterance cannot be
split further into a smaller component subutterance and word, that is,
when its length is zero.  Suppose for example, that a given utterance
is {\em abcde}, where the letters represent phonemes.  If $\seg(x)$
represents the best segmentation of the utterance $x$ and $\word(x)$
denotes that $x$ is treated as a word, then
$$
\seg(abcde) = {\bf best \,\, of\,\,} \left\{\begin{array}{l}
	\word(abcde) \\
	\seg(a) + \word(bcde)\\
	\seg(ab) + \word(cde)\\
	\seg(abc) + \word(de)\\
	\seg(abcd) + \word(e) \end{array} 
	\right.
$$
The {\bf evalUtterance} algorithm in Figure~\ref{fig:dyn-rec} does
precisely this.  It initially assumes the entire input utterance to be
a word on its own by assuming a single segmentation point at its right
end. It then compares the log probability of this segmentation
successively to the log probabilities of segmenting it into all
possible subutterance, word pairs.  Once the best segmentation into
words has been found, then spaces are inserted into the utterance at
the inferred points and the segmented utterance is printed out.

The implementation maintains two separate tables internally, one for
words and one for phonemes.  When the procedure is initially started,
the word table is empty.  Only the phoneme table is populated with
equipossible phonemes.  As the program considers each utterance in
turn and commits to its best segmentation according to the {\bf
evalUtterance} algorithm, the two tables are updated correspondingly.
For example, after some utterance ``abcde'' is segmented into ``a bc
de'', the word table is updated to increment the frequencies of the
three entries ``a'', ``bc'' and ``de'' each by 1, and the phoneme table
is updated to increment the frequencies of each of the phonemes in the
utterance including one sentinel for each word inferred.  Of course,
incrementing the frequency of a currently unknown word is equivalent
to creating a new entry for it with frequency 1.

\begin{figure}[htb]
\begin{small}
\subsection{Algorithm: evalUtterance}
\begin{verbatim}
BEGIN
   Input (by ref) utterance u[0..n]
   where u[i] are the characters in it.

   bestSegpoint := n;
   bestScore := evalWord(u[0..n]);
   for i from 0 to n-1; do
      subUtterance := copy(u[0..i]);
      word := copy(u[i+1..n]);
      score := evalUtterance(subUtterance)
               + evalWord(word);
      if (score < bestScore); then
         bestScore = score;
         bestSegpoint := i;
      fi
   done
   insertWordBoundary(u, bestSegpoint)
   return bestScore;
END
\end{verbatim}
\end{small}
\caption{Recursive optimization algorithm to find the best
segmentation of an input utterance using the language model
described in this paper.}
\label{fig:dyn-rec}
\end{figure}

\begin{figure}[htb]
\begin{small}
\subsection{Function: evalWord}
\begin{verbatim}
BEGIN
   Input (by reference) word w[0..k]
   where w[i] are the phonemes in it.

   score := 0;
   N := number of distinct words;
   S := sum of their frequencies;
   if freq(word) == 0; then {
      escape := N/(N+S);
      P_0 := relativeFrequency('#');
      score := -log(esc) -log(P_0/(1-P_0));
      for each w[i]; do
         score -= log(relativeFrequency(w[i]));
      done
   } else {
      P_w := frequency(w)/(N + S);
      score := -log(P_w);
   }
   return score;
END
\end{verbatim}
\end{small}
\caption{The function to compute $-\log \p(w)$ of an input word $w$.
If the word is novel, then the function backs off to using a
distribution over the phonemes in the word.}
\label{fig:word-len}
\end{figure}

One can easily see that the running time of the program is $O(mn^2)$
in the total number of utterances ($m$) and the length of each
utterance ($n$), assuming an efficient implementation of a hash table
allowing nearly constant lookup time is available.  A single run
over the entire corpus typically completes in under 10 seconds on a
300 MHz i686-based PC running Linux 2.2.5-15.  Although all the discussed
algorithms tend to complete within one minute on the reported corpus,
MBDP-1's running time is quadratic in the number of utterances, while
the language model presented here enables computation in almost linear
time.  The typical running time of MBDP-1 on the 9790-utterance corpus
averages around 40 seconds per run on a 300 MHz i686 PC while the
algorithm described in this paper averages around 7 seconds.

\section{Results and Discussion}
\label{sec:results}

\begin{figure}[htb]
\begin{center}
  \includegraphics[width=3.25in]{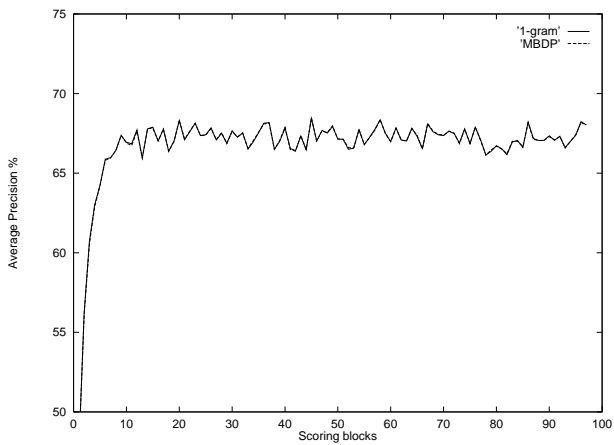} 
  \caption{Word discovery precision as a function of number of
    utterances considered.  Each scoring block (checkpoint) consists of
    10\% of the total number of utterances (roughly 1000).  It is hard
    to discern two separate plots above because of the close match in
    their performance.  1-gram denotes the performance of the
    procedure reported in this paper whereas MBDP denotes the
    performance of Brent's Model Based Dynamic Programming algorithm.
  }
  \label{fig:r-pre}
\end{center}
\end{figure}

\begin{figure}[htb]
\begin{center}
  \includegraphics[width=3.25in]{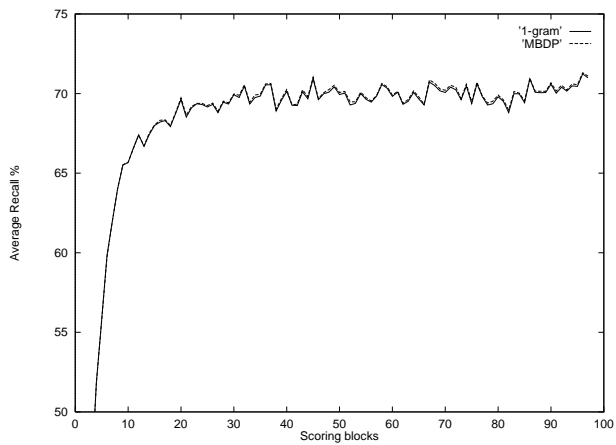}
  \caption{Word discovery recall as a function of number of utterances considered.}
  \label{fig:r-rec}
\end{center}
\end{figure}

\begin{figure}[htb]
\begin{center}
  \includegraphics[width=3.25in]{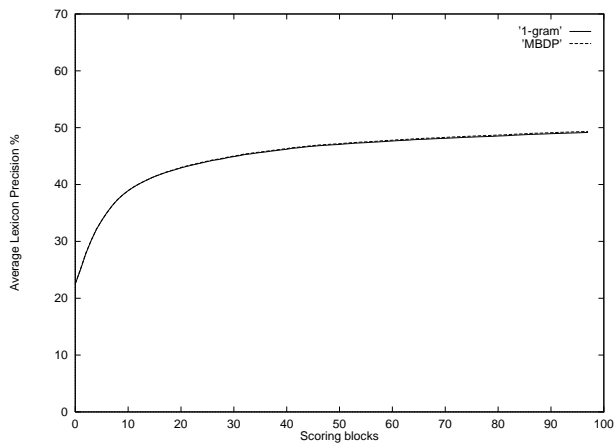}
  \caption{Lexicon precision (percentage of correctly inferred words
  in the lexicon) as a function of number of utterances considered.
  }
  \label{fig:r-lex}
\end{center}
\end{figure}

In line with the results reported in \emcite{Brent:EPS99}, three
scores were calculated --- precision, recall and lexicon precision.
Precision is defined as the proportion of predicted words that are
actually correct.  Recall is defined as the proportion of correct
words that were predicted.  Lexicon precision is defined as the
proportion of words in the predicted lexicon that are correct.
Precision and recall scores were computed incrementally and
cumulatively within scoring blocks, each of which consisted of 100
utterances.  We emphasize that the segmentation itself proceeded
incrementally, on an utterance-by-utterance basis.  Only the scores
are reported on a per-block basis for brevity.  These scores were
computed and averaged only for the utterances within each block scored
and thus they represent the performance of the algorithm on the block
of utterances scored, occurring in the exact context among the other
scoring blocks.  Lexicon scores carried over blocks cumulatively.  As
Figures~\ref{fig:r-pre} through~\ref{fig:r-lex} show, the performance
of our algorithm matches that of MBDP-1 on all grounds.  In fact, we
found to our surprise that the performances of both algorithms were
almost identical except in a few instances, discussion of which space
does not permit here.
%----------------------------------------------------------------------

%----------------------------------------------------------------------

This leads us to suspect the two, substantially different, statistical
models may essentially be capturing the same nuances of the domain.
Although \emcite{Brent:EPS99} explicitly states that probabilities are
not estimated for words, it turns out that considering the entire
corpus as a single event in probability space does end up having the
same effect as estimating probabilities from relative frequencies as
our statistical model does.  The {\em relative probability}\/ of a
familiar word is given in Equation~22 of \emcite{Brent:EPS99} as
$$
\frac{f_k(\hat{k})}{k}\cdot
\left( \frac{f_k(\hat{k})-1}{f_k(\hat{k})}\right)^2
$$
where $k$ is the total number of words and $f_k(\hat{k})$ is the
frequency at that point in segmentation of the $\hat{k}$th word. It 
effectively approximates to the relative frequency 
$$
\frac{f_k(\hat{k})}{k}
$$
as $f_k(\hat{k})$ grows.  The language model presented in this paper
explicitly claims to use this specific estimator for the word
probabilities.  From this perspective, both MBDP-1 and the present
model tend to favor the segmenting out of familiar words that do not
overlap.  In this context, we are curious to see how the algorithms
would fare if in fact the utterances were favorably ordered, that is,
in order of increasing length.  Clearly, this is an important
advantage for both algorithms.  The results of experimenting with a
generalization of this situation, where instead of ordering the
utterances favorably, we treat an initial portion of the corpus as a
training component effectively giving the algorithms free word
boundaries after each word, are presented in
Section~\ref{sec:training}.

In contrast with MDBP-1, we note that the model proposed in this paper
has been entirely developed along conventional lines and has not made
the somewhat radical assumption of treating the entire observed corpus
as a single event in probability space.  Assuming that the corpus
consists of a single event requires the explicit calculation of the
probability of the lexicon in order to calculate the probability of
any single segmentation.  This calculation is a nontrivial task since
one has to sum over all possible orders of words in the lexicon,
$\Lex$.  This fact is recognized in \emcite{Brent:EPS99}, where the
expression for $\p(\Lex)$ is derived in Appendix~1 of his paper as an
approximation.  One can imagine then that it will be correspondingly
more difficult to extend the language model in \emcite{Brent:EPS99}
past the case of unigrams.  As a practical issue, recalculating
lexicon probabilities before each segmentation also increases the
running time of an implementation of the algorithm.

Furthermore, the language model presented in this paper estimates
probabilities as relative frequencies using the commonly used back-off
procedure and so they do not assume any priors over integers.
However, MBDP-1 requires the assumption of two distributions over
integers, one to pick a number for the size of the lexicon and another
to pick a frequency for each word in the lexicon.  Each is assumed
such that the probability of a given integer $\p(i)$ is given by
$\frac{6}{\pi^2i^2}$.  We have since found some evidence suggesting
that the choice of a particular prior does not have any significant
advantage over the choice of any other prior.  For example, we have
tried running MBDP-1 using $\p(i)=2^{-i}$ and still obtained
comparable results.  It is noteworthy, however, that no such
subjective prior needs to be chosen in the model presented in this
paper.

The other important difference between MBDP-1 and the present model is
that MBDP-1 assumes a uniform distribution over all possible word
orders and explicitly derives the probability expression for any
particular ordering.  That is, in a corpus that contains $n_k$
distinct words such that the frequency in the corpus of the $i$th
distinct word is given by $f_k(i)$, the probability of any one
ordering of the words in the corpus is
$$
\frac{\prod_{i=1}^{n_k} f_k(i)!}{k!}
$$ 
because the number of unique orderings is precisely the reciprocal of
the above quantity.  In contrast, this independence assumption is
already implicit in the unigram language model adopted in the present
approach.  Brent mentions that there may well be efficient ways of
using $n$-gram distributions within MBDP-1.  However, the framework
presented in this paper is a formal statement of a model that lends
itself to such easy $n$-gram extensibility using the back-off scheme
proposed.  It is now a simple matter to include bigrams and trigrams
among the tables being learned.  Since back-off has already been
incorporated into the model, we simply substitute for the probability
expression of a word (which currently uses no history), the
probability expression given its immediate history (typically $n-1$
words).  Thus, we use an expression like
\begin{eqnarray*}
\p(w|h) &=& \left\{
	\begin{array}{ll}
	\alpha \frac{C(h,w)}{C(h)} &
	{\rm if\ } C(h,w) > 0 \\
	(1-\alpha) \p(w|h') & {\rm otherwise}\\
	\end{array}
	\right.
\end{eqnarray*}
where $\p(w|h)$ denotes the probability of word $w$ conditioned on its
history $h$, normally the immediately previous 1 (for bigrams) or 2
(for trigram) words, $\alpha$ is the back-off weight or discount
factor, which we may calculate using any of a number of standard
techniques, for example by using the Witten-Bell technique as we have
done in this paper, $C()$ denotes the count or frequency function of
its argument in its respective table, and $h'$ denotes reduced history,
usually by one word.  Reports of experiments with such extensions can,
in fact, be found in a forthcoming article \cite{Venkataraman:ASM01}.

\section{Training}
\label{sec:training}

\begin{figure}[htb]
\begin{center}
  \includegraphics[width=3.25in]{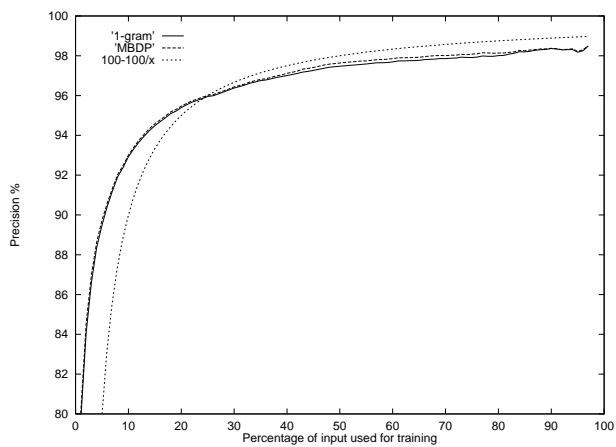}
  \caption{Responsiveness of the algorithm to training information.
  The horizontal axis represents the initial percentage of the data
  corpus that was used for training the algorithm.  This graph shows
  the improvement in segmentation precision with training size.}
  \label{fig:t-pre}
\end{center}
\end{figure}

\begin{figure}[htb]
\begin{center}
  \includegraphics[width=3.25in]{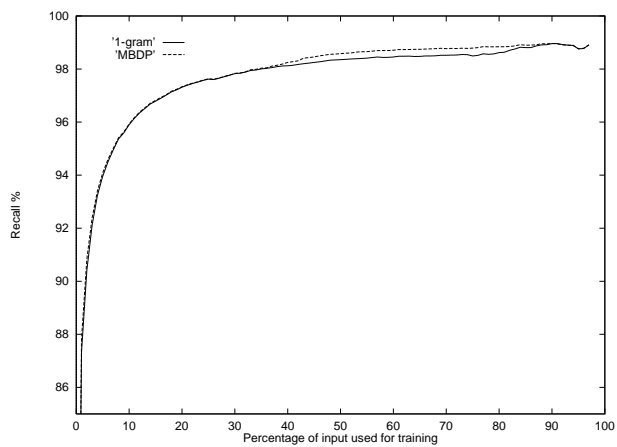}
  \caption{Improvement in segmentation recall with training size.}
  \label{fig:t-rec}
\end{center}
\end{figure}

Although we have presented the algorithm as an unsupervised learner,
it is interesting to compare its responsiveness to the effect of
training data.  Here, we extend the work in \emcite{Brent:EPS99} by
reporting the effect of training upon the performance of both
algorithms.  Figures~\ref{fig:t-pre} and~\ref{fig:t-rec} plot the
results (precision and recall) over the whole input corpus, that is,
blocksize = $\infty$, as a function of the initial proportion of the
corpus reserved for training.  This is done by dividing the corpus
into two segments, with an initial training segment being used by the
algorithm to learn word and phoneme probabilities and the latter
actually being used as the test data.  A consequence of this is that
the amount of data available for testing becomes progressively smaller
as the percentage reserved for training grows.  So, the significance of
the test would diminish correspondingly.  We may assume that the plots
cease to be meaningful and interpretable when more than about 75\%
(about 7500 utterances) of the corpus is used for training.  At 0\%,
there is no training information for any algorithm, and the
performances of the various algorithms are identical to those of the
unsupervised case.  We increase the amount of training data in steps
of approximately 1\% (100 utterances).  For each training set size,
the results reported are averaged over 25 runs of the experiment, each
over a separate random permutation of the corpus.  The motivation was
both to account for ordering idiosyncrasies and to smooth the graphs
to make them easier to interpret.

%----------------------------------------------------------------------

%----------------------------------------------------------------------

We interpret Figures~\ref{fig:t-pre} and~\ref{fig:t-rec} as suggesting
that the performance of all the discussed algorithms can be boosted
significantly with even a small amount of training.  It is also
noteworthy and reassuring to see that, as one would expect from
results in computational learning theory \cite{Haussler:QIB88}, the
number of training examples required to obtain a desired value of
precision, $p$, appears to grow with $1/(1-p)$.

\subsection*{Significance of single word utterances}

The results we have obtained provide some insight into the actual
learning process, which appears to be one in which rapid bootstrapping
happens with very limited data.  As we had remarked earlier, all the
internal tables are initially empty.  Thus, the very first utterance is
necessarily segmented as a single novel word.  The reason that fewer
novel words are preferred initially is this: Since the word table is
empty when the algorithm attempts to segment the first utterance,
backing-off causes all probabilities to necessarily be computed from
the level of phonemes up.  Thus, the more words in it, the more
sentinel characters that will be included in the probability
calculation and so that much lesser will be the corresponding
segmentation probability.  As the program works its way through the
corpus, correctly inferred words, by virtue of their relatively
greater preponderance compared to noise, tend to dominate the
distributions and thus dictate how future utterances are segmented.

From this point of view, we see that the presence of single word
utterances is of paramount importance to the algorithm.  Fortunately,
very few such utterances suffice for good performance, for every
correctly inferred word helps in the inference of other words that are
adjacent to it.  This is the role played by {\em training}, whose
primary use can now be said to be in supplying the word table with
{\em seed}\/ words.  We can now further refine our statement about the
importance of single word utterances.  Although single word utterances
are important for the learning task, what are critically important are
words that occur both by themselves in an utterance and in the context
of other words after they are first seen.  This brings up a
potentially interesting issue.  Suppose disfluencies in speech can be
interpreted, in some sense, as {\em free word boundaries}.  We are
then interested in whether their distribution in speech is high enough
in the vicinity of generally frequent words.  If that is the case,
then {\em um}s and {\em ah}s are potentially useful from a cognitive
point of view to a person acquiring a lexicon since these are the very
events that will bootstrap his or her lexicon with the initial seed
words that are instrumental in the rapid acquisition of further words.

\section{Summary}
\label{sec:summary}

In summary, we have presented a formal model of word discovery in
speech transcriptions.  The main advantages of this model over that of
\emcite{Brent:EPS99} are, first, that the present model has been
developed entirely by direct application of standard techniques and
procedures in speech processing.  Second, the model is easily
extensible to incorporate more historical detail in the usual way.
Third, the presented model makes few assumptions about the nature of
the domain and remains as far as possible conservative and simple in
its development.  Results from experiments suggest that the algorithm
performs competitively with other unsupervised techniques recently
proposed for inferring words from transcribed speech.  Finally,
although the algorithm is originally presented as an unsupervised
learner, we have shown the effect that training data has on its
performance.

\subsection*{Future work}

Other extensions being worked on include the incorporation of more
complex phoneme distributions into the model.  These are, namely, the
biphone and triphone models.  Using the lead from
\emcite{Brent:EPS99}, attempts to model more complex distributions for
words such as those based on {\em template grammars\/} and the
systematic incorporation of prosodic, stress and phonotactic
constraint information into the model are also the subject of current
interest.  We already have some unpublished results suggesting that
biasing the segmentation using a constraint that every word must have
at least one vowel in it dramatically increases segmentation precision
from 67.7\% to 81.8\%, and imposing a constraint that words can begin
or end only with permitted clusters of consonants increases precision
to 80.65\%.

Another avenue of current research is concerned with iterative
sharpening of the language model wherein word probabilities are
periodically reestimated using a fixed number of iterations of the
Expectation Modification (EM) algorithm \cite{Dempster:MLF77}.  Such
reestimation has been found to improve the performance of language
models in other similar tasks.  It has also been suggested that the
algorithm could be usefully adapted to user modeling in human-computer
interaction, where the task lies in predicting the most likely {\em
atomic}\/ action a computer user will perform next.  However, we have
as yet no results or work to report on in this area.

\section{Acknowledgments} 
 
The author thanks Michael Brent for stimulating his interest in the
area.  Thanks are also due to Koryn Grant for cross-checking the
results presented here.  Eleanor Olds Batchelder and Andreas Stolcke
offered many constructive comments and useful pointers in preparing a
revised version of this paper.  Anonymous reviewers of an initial
version helped significantly in improving its content and Judy Lee
proof-read the final version carefully.

\bibliography{seg}

\end{document}